\documentclass[10pt,twocolumn,letterpaper]{article}

\usepackage[pagenumbers]{cvpr} 

\makeatletter
\@namedef{ver@everyshi.sty}{}
\makeatother

\usepackage{graphicx}
\usepackage{amsmath}
\usepackage{amssymb}
\usepackage{booktabs}

\usepackage[utf8]{inputenc} 
\usepackage[T1]{fontenc}    

\usepackage{url}            
\usepackage{amsfonts}       
\usepackage{nicefrac}       
\usepackage{microtype}      
\usepackage{comment}
\usepackage{graphicx}
\usepackage{morenotations}
\DeclareMathOperator*{\argmax}{arg\,max}

\newcommand{\alan}[1][blue]{\textcolor{#1}}

\def\cvprPaperID{5392} 
\def\confName{CVPR}
\def\confYear{2023}

\begin{document}

\title{Knowledge Combination to Learn Rotated Detection Without Rotated Annotation}

\author{
Tianyu Zhu\textsuperscript{\rm 1, 2,}\thanks{Corresponding author.} \quad
Bryce Ferenczi\textsuperscript{\rm 2} \quad
Pulak Purkait\textsuperscript{\rm 1} \quad
Tom Drummond\textsuperscript{\rm 3} \quad \\
Hamid Rezatofighi\textsuperscript{\rm 2} \quad
Anton van den Hengel\textsuperscript{\rm 1} \\
\textsuperscript{\rm 1}Amazon IML. \quad
\textsuperscript{\rm 2}Monash University \quad
\textsuperscript{\rm 3}The University of Melbourne \\
{\tt\small tianyuzhu52@gmail.com} \quad
{\tt\small\{alanzty, purkaitp, hengelah\}@amazon.com} \quad
{\tt\small tom.drummond@unimelb.edu.au} \\
{\tt\small\{tianyu.zhu, bryce.ferenczi, hamid.rezatofighi\}@monash.edu}
}

\maketitle

\begin{abstract}
Rotated bounding boxes drastically reduce output ambiguity of elongated objects, making it superior to axis-aligned bounding boxes. 
Despite the effectiveness, rotated detectors are not widely employed. Annotating rotated bounding boxes is such a laborious process that they are not provided in many detection datasets where axis-aligned annotations are used instead. 
In this paper, we propose a framework that allows the model to predict precise rotated boxes only requiring cheaper axis-aligned annotation of the target dataset~\footnote{Code is available at: 
{https://github.com/alanzty/KCR-Official}}. 

To achieve this, we leverage the fact that neural networks are capable of learning richer representation of the target domain than what is utilized by the task. The under-utilized representation can be exploited to address a more detailed task. Our framework combines task knowledge of an out-of-domain source dataset with stronger annotation and domain knowledge of the target dataset with weaker annotation. A novel assignment process and projection loss are used to enable the co-training on the source and target datasets. As a result, the model is able to solve the more detailed task in the target domain, without additional computation overhead during inference.
We extensively evaluate the method on various target datasets including fresh-produce dataset, HRSC2016 and SSDD. Results show that the proposed method consistently performs on par with the fully supervised approach.

\noindent\textbf{Acknowledgement} This paper is inspired by a computer vision project conducted at Amazon. We would like to express our sincere gratitude to the following individuals for their contributions to this research project: Gil Avraham, Hisham Husain, Chenchen Xu, Ravi Garg, Shatanjay Khandelwal and Philip Schulz, who all work at Amazon. Their support, insights, and feedback were invaluable throughout the research process, and we are truly grateful for their help.

\end{abstract}

\section{Introduction}
Rotated detectors introduced in recent works~\cite{xia2018dota, liu2017high, nayef2017icdar2017} have received attention due to their outstanding performance for top view images~\cite{xie2021oriented, yang2021learning, liu2017rotated}. They reduce the output ambiguity of elongated objects for downstream tasks making them superior to axis-aligned detectors in dense scenes with severe occlusions~\cite{looi2019rotatedmrcnn}. However, the rotated annotation is more expensive compared to axis-aligned annotation. Furthermore, popular 2D annotation tools such as Sagemaker Groundtruth~\footnote{{https://aws.amazon.com/sagemaker/data-labeling/}} and VGG app~\footnote{{https://www.robots.ox.ac.uk/vgg/software/via/}} do not support rotated bounding box annotations. As a result, many popular detection datasets only have axis-aligned annotations~\cite{kuznetsova2020open, mot20, everingham2010pascal}. These problems reduces the potential scope of the implementation of rotated detectors. In this work, we introduce \textbf{K}nowledge \textbf{C}ombination to learn \textbf{R}otated object detection, a training scheme that only requires cheaper axis-aligned annotation for the target dataset in order to predict rotated boxes. 

\begin{figure}
  \centering
  \includegraphics[width=\linewidth]{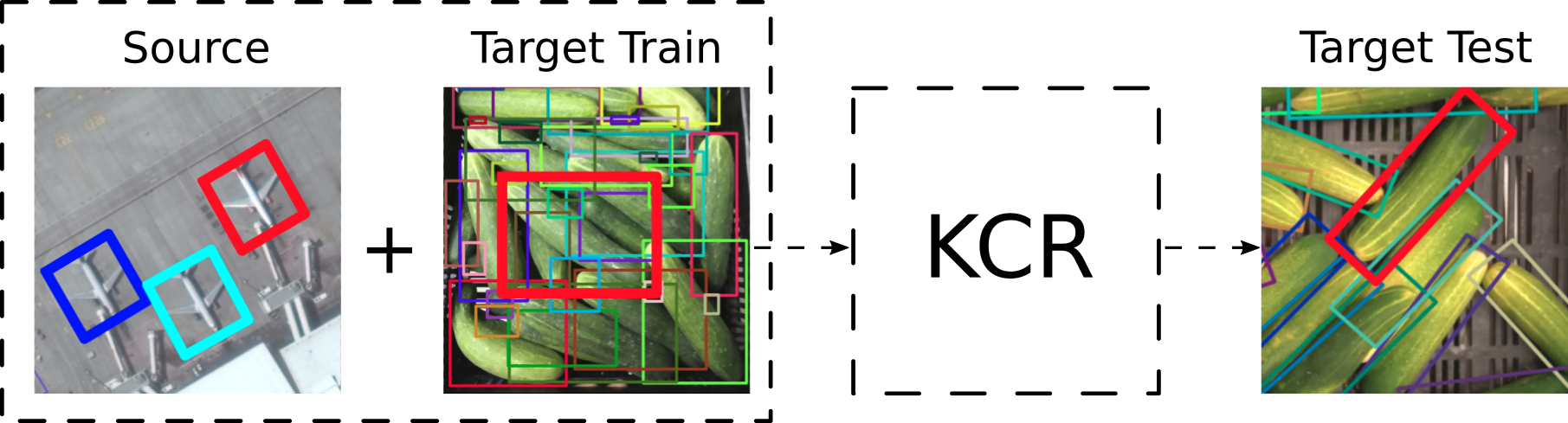}
  \caption{KCR combines the task knowledge of a source dataset with stronger rotated annotation and the domain knowledge of the target dataset with weaker axis-aligned annotation, which enables the model to predict rotated detection on the target domain.}
  \label{Figure: teaser}
\end{figure}

Neural networks encode data into a latent space, which is then decoded to optimize the given task. The latent embedding is an abstract representation of the data, containing much richer information than the output~\cite{tishby2015deep}. Early works in deep learning show that the model implicitly learns to detect image features such as edges and corners~\cite{krizhevsky2017imagenet, lecun1998gradient}, which can be used for more detailed tasks if decoded properly. 
We believe decoding to a more precise task on the target domain can be learnt via co-optimizing with a strongly labelled source dataset.
We design a framework that combines task knowledge of rotated detection from a source dataset, and the domain knowledge of a disjoint class of objects in the target dataset with only axis-aligned annotation, as shown in Figure~\ref{Figure: teaser}. This approach combines the advantage of both weakly-supervised learning and transfer learning.

We follow a design principal that the framework should maximize the target domain knowledge learnt by the model while minimizing the negative impact caused by weaker labels. 
This is achieved by co-training the source and target dataset with projection losses and a novel assignment process. The design choices are validated through ablation studies. We conduct extensive experiments to demonstrate that our framework is robust to a large domain gap between source and target dataset. 
Therefore, box orientation can practically be learnt for free with KCR, due to the availability of free public source datasets such as DOTA~\cite{xia2018dota} with rotated annotations.
We show the efficacy of this method on a fresh-produce dataset with high density of objects and severe occlusions. The performance (AP50) gap between the proposed method, learning from weak axis-aligned boxes, and the fully-supervised model learning from strong rotated annotation, reduces to only $3.2\%$ for the challenging cucumber dataset.
We apply the same framework to HRSC2016~\cite{liu2016ship} and SSDD~\cite{SSDD2021} datasets to show that our method consistently performs on par with fully supervised models. The performance gap reduces to $1.0\%$ for SSDD.
We believe our approach can greatly increase the usage and impact of rotated object detectors. The source code will be publicly available for the community to save future annotation cost.
In summary, our main contributions are as follows:
\setlist{nolistsep}
\begin{enumerate}
	\setlength{\itemsep}{1pt}
	\setlength{\parskip}{0pt}
	\setlength{\parsep}{0pt}
    \item[1)] We introduce a framework that combines task knowledge of a strongly labelled source dataset and domain knowledge of a weakly labelled target dataset.  
    \item[2)] We apply this method in 2D rotated detection task, enabling the model to predict rotated bounding box with only axis-aligned annotation and verify the generality of the method with several datasets.
    \item[3)] We demonstrate robustness of the framework to various domain gaps between source and target datasets. Hence, box orientation can be learnt with no additional annotation cost in practical applications.
\end{enumerate}
\section{Related Work}

\noindent\textbf{Rotated Detection Task.}
Rotated object detection requires the model to predict minimum area rectangles with five degrees of freedom, namely rotated bounding boxes, enclosing objects of interests~\cite{xia2018dota}. In axis-aligned object detection, the output rectangles have four degrees of freedom, which are aligned with the image  axes~\cite{lin2014coco}.
The rotated boxes occupy a much smaller area when estimating the location status of diagonally positioned elongated objects as shown in Figure~\ref{Figure: teaser}. Rotated detection is strictly superior to axis-aligned detection as there is less background within the box, and the orientation can potentially convey object pose information. However, there are only a handful of datasets with rotated annotations~\cite{xia2018dota, liu2016ship, SSDD2021} comparing to large number of readily available large-scale axis-aligned datasets~\cite{kuznetsova2020open, mot17, everingham2010pascal, lin2014coco}. A potential contributor to this phenomenon is the ease of annotating axis-aligned boxes by simple click-and-drag with current labelling tools. Rotated boxes require much more effort to tightly enclose the objects with an extra degree of freedom. Popular annotation tools such as AWS Sagemaker and VGG app do not support rotated boxes. In order to acquire tighter rotated annotations, users must pay for instance segmentation, which is significantly more expensive and unnecessary for the final task. Therefore, we propose this work to address the shortcomings of the rotated object detection task pipeline by making orientation free to learn. 

\noindent \textbf{Weakly-Supervised Learning.}
Weakly supervised learning sits between fully supervised learning and unsupervised learning, in the sense that only weak labels are available. The labels are weak either because they are incomplete, inexact or noisy~\cite{gokberk2014multi, ren2020instance, wei2018revisiting, kolesnikov2016seed, ahn2018learning, wang2020self}.
In computer vision, popular weakly-supervised learning tasks include object detection with only image level annotation~\cite{feng2022weakly, zhang2021weakly} and instance segmentation with only box annotation~\cite{papandreou2015weakly, dai2015boxsup}. Due to the difficulty of pixel-wise prediction, weakly-supervised instance segmentation still falls significantly behind a fully-supervised model~\cite{tian2021boxinst} on novel objects. 
In this paper, we focus on learning rotated detection requiring five parameters with only axis-aligned annotation during training which provides four parameters. The one parameter difference makes our approach weakly-supervised.
To the best our knowledge, such problem has only been attempted on specific category of objects~\cite{iqbal2021leveraging} but not approached generally. 
For elongated objects, solving this problem is more appropriate than solving weakly-supervised instance segmentation directly.

\begin{figure*}
  \centering
  \includegraphics[width=0.9\linewidth]{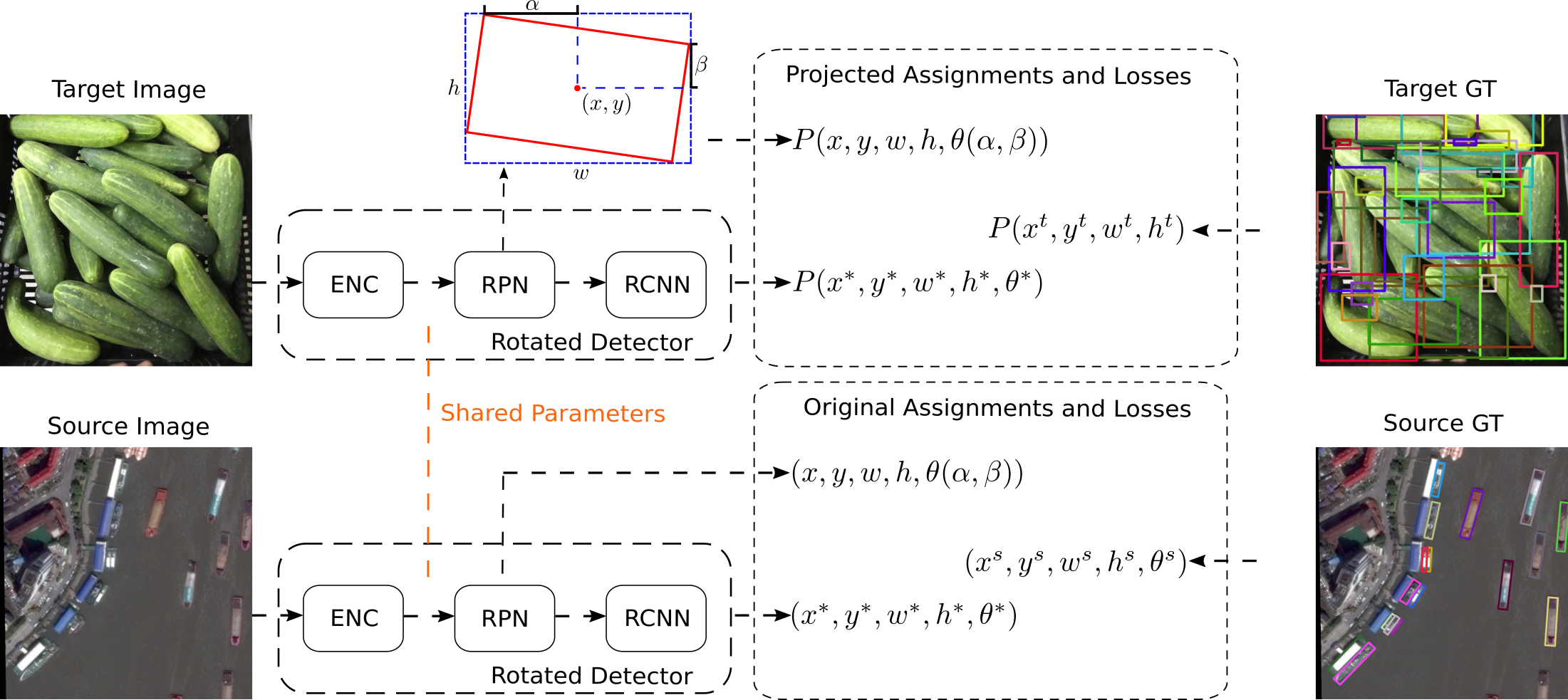}
  \caption{Overall framework of KCR, which learns the knowledge of rotated detection from source dataset combined with domain knowledge of axis-aligned target dataset to infer rotated bounding boxes of target objects. The rotated detector~\cite{xie2021oriented} takes in an image encoded by a CNN and generates \textit{first-stage} proposals $(x, y, w, h, \alpha ,\beta, p)$ and \textit{second-stage} refinements $(x^*, y^*, w^*, h^*, \theta^*, p^*)$. We use a function $P$ to either project or enlarge the box representation during the assignment processes and loss functions.}
  \label{Figure: dota12cucumber}
\end{figure*}

\noindent \textbf{Transfer Learning.}
Computer vision models are frequently initialised with backbones\cite{he2016deep, redmon2013darknet, liu2021swin} pretrained on ImageNet or COCO~\cite{lin2014coco, russakovsky2015imagenet}, which is a basic form of transfer learning and a common practice. 
Transfer learning is highly effective when the target dataset has small sample size such as medical imaging~\cite{ronneberger2015u}. 
Despite the fact that conventional transfer learning reduces the number of data samples required for a specific task, strong annotation of the target domain is still required for the model to learn the detailed task. In this paper, we utilize co-training strategy to transfer the ability to make more detailed predictions from the source to the target dataset with weaker annotations.

\section{KCR}
The goal of our work is to learn to predict rotated bounding boxes on a \textit{target} dataset of which we only have axis-aligned bounding boxes in training examples. We develop a co-training scheme that utilizes an out-of-domain but strongly labelled \textit{source} dataset to learn accurate rotations of elongated objects. 
An example of a potential source and target dataset pair are a satellite imagery dataset, DOTA~\cite{xia2018dota}, and a fresh-produce dataset, as shown in Figure~\ref{Figure: teaser}. In this section, we first briefly describe the workflow of the detector following by our training scheme which enables weakly-supervised learning and knowledge transfer. 

\subsection{Rotated Detection Overview}
In this work, we employ the same architectural choices as~\cite{xie2021oriented} and the flow is briefly demonstrated in Figure~\ref{Figure: dota12cucumber}. The forward propagation comes in two stages: the oriented RPN followed by oriented R-CNN, where both methods contain a classification head and a regression head. The RPN takes an image and generates $N$ oriented region proposals $\braces{R_i}_{i=1}^N$ which each take the form $R_i = (x_i,y_i,w_i,h_i,\alpha, \beta, p_i)$, where $(x_i,y_i)$ denotes the center, $w_i$ and $h_i$ are the width and height of the tightest axis-aligned external box. $\alpha$ and $\beta$ are the are the offsets relative to the midpoints of the top and right sides of the external rectangle. $p_i$ is an object score. The proposed rotated region will then be cropped by rotated roi~\cite{xie2021oriented} in feature space. The proposal is then fed to Oriented R-CNN which is a CNN followed by another classification head and a bounding box regression head rectifying the spatial location. We denote the output of of second stage is $\braces{R_i^*}_{i=1}^N$ and $R_i^* = (x_i^*,y_i^*,w_i^*,h_i^*, \theta_i^*, p_i^*, c_i^*)$ where $(x_i^*, y_i^*)$ denotes the center, $w_i^*$ and $h_i^*$ are the width and height and $\theta_i^*$ the rotation angle of the final predicted rotated box. $p_i^*$ is the second stage object score and $c_i^*$ represents the classification score. 

\subsection{Learning Rotated Region Proposal}
\label{ssec:oriented_pred}
We develop a co-training scheme that utilizes both source and target datasets. We denote $\braces{B_j^s}_{j=1}^m$ to be the $m$ rotated bounding boxes of a source image where $B_j^s = (x_j^s,y_j^s,w_j^s,h_j^s, \theta_j^s)$ is a rotated box. For the target dataset, we have $n$ axis-aligned boxes $\braces{B_j^t}_{j=1}^n$ on a target image where $B_j^t = (x_j^t,y_j^t,w_j^t,h_j^t)$ is a axis-aligned box.

In the first stage, a set of $N$ oriented regions are proposed where each proposal $R_i = (x_i,y_i,w_i,h_i,\alpha, \beta, p_i)$ is assigned a ground truth label $B_{\sigma(i)}$ based on intersection over union (IoU) matching, where the assignment $\sigma(i)$ and matching score $\tau(i)$ are 
\begin{align}
    \sigma(i) = \argmax_{j \in \braces{1,\ldots,m}} iou((x_i,y_i,w_i,h_i), P(B_j)), \\
    \tau(i) = \max_{j \in \braces{1,\ldots,m}} iou((x_i,y_i,w_i,h_i), P(B_j)).
\end{align}
Since in first stage, $(x_i,y_i,w_i,h_i)$ represents the tightest external axis-aligned box instead of the rotated region itself, we need a transformation function $P$ to project the ground truth to axis-aligned box for source rotated dataset. It is important to notice that $P(B_j)$ is strictly equal or larger than canonical axis-aligned box of the object. Therefore, for target axis-aligned dataset, we need to enlarge $B_j^t$. Here we formally define the transformation function

\begin{align}
    P(B) = 
    \begin{cases}
        (x_{min}, y_{min}, x_{max}, y_{max}) \\ 
        ~~~~~~~~~~~~~~\text{ for $B = (x, y, w, h, \theta)$, }\\
        (x_{min} - \gamma w_i, y_{min} - \gamma h_i, x_{max} + \gamma w_i, \\
        ~~~~~~~~~~~~~~y_{max} + \gamma h_i) \text{ for $B = (x, y, w, h)$, }
    \end{cases}
\end{align}
where $\gamma$ is simply an enlargement factor that we can tune.  
The loss for this RPN is defined as follows
\begin{align}
    L_{\mathcal{S}} & = \frac{1}{N} \sum_{i=1}^N -\mathbf{1}_{{\tau(i)} \geq 0.5}\log(p_i) + l_1(R_i, B_{\sigma(i)}^s)
\end{align}
\vspace{-2em}
\begin{multline*}
    ~~L_{\mathcal{T}} = \frac{1}{N} \sum_{i=1}^N -\mathbf{1}_{{\tau(i)} \geq 0.5}\log(p_i) \\ + l_1((x_i,y_i,w_i,h_i), P(B_{\sigma(i)}^t)),
\end{multline*}
\begin{align}    
    L &= L_{\mathcal{S}} + L_{\mathcal{T}},~~~~~~~~~~~~~~~~~~~~~~~~~~~~~~~~~~~~~~~~~~~~~~~~~~~~~~~~
    \label{eq: loss rpn}
\end{align}      
where the loss for both target and source examples are composed by a binary cross entropy (BCE) loss and an $l_1$ regression loss on spatial outputs. The label for the BCE is determined by whether the proposal has an overlap larger than 0.5 with any projected ground truth. For the source dataset, we can compute ground truth $(\alpha^s, \beta^s)$ from $\theta^s$ which is the regression for the rotation representation. However, for the target dataset, we only compute regression loss for $(x_i,y_i,w_i,h_i)$. The model must learn the rotation knowledge from the source dataset. To simplify the mathematical notation, we omit classification loss as it is less relevant to our contribution than object score.

\subsection{Learning Rotated R-CNN}
\label{subsec: Learning Rotated R-CNN}
In the second stage, the Oriented R-CNN takes in a subset of proposals generated in first stage and make final prediction $R_i^* = (x_i^*,y_i^*,w_i^*,h_i^*, \theta_i^*, p_i^*, c_i^*)$. The bounding box regression is less important because of two reasons. Firstly, the cropped region proposal is already an approximate detection. The goal of regression of this stage is to fine-tune. Secondly, we choose to use class-agnostic rotated bounding box regression here as the model is able to learn general bounding box regression from the source dataset. 
The classification of the second stage is also straight forward to train because canonical axis-aligned detector fundamentally identical in that respect. 
The most important and challenging aspect is to produce an accurate object score $p_i^*$, which is a direct result of ground truth assignment process. We first formulate the assignment process for the source dataset as

\begin{align}
    \sigma_s^*(i) = \argmax_{j \in \braces{1,\ldots,m}} iou((x_i,y_i,w_i,h_i, \theta_i), B_j^s), \\
    \tau_s^*(i) = \max_{j \in \braces{1,\ldots,m}} iou((x_i,y_i,w_i,h_i, \theta_i), B_j^s).
    \label{eq: rcnn source}
\end{align}
Note that the assignment process is based on the proposal instead of the final refined prediction. The difference from the first stage is that here we use accurate rotated ground truth $B_j^s$ against $R_i$ instead of the external enclosing axis-aligned box. We compute $\theta$ of the first stage with $(\alpha, \beta)$ inline with ~\cite{xie2021oriented}. 
This works appropriately with the source dataset. However, it creates a challenge for the target dataset which lacks rotation information. If we follow the identical assignment process as the source dataset, then a close-to-correct proposal might be classified as negative example if the overlap between the rotated bounding box and axis-aligned bounding box is low. Consequently, false negative rate will be increased, reducing the recall of such training scheme. Therefore, we propose two additional strategies, projection assignment and heuristic selection.

\noindent\textbf{Projection Assignment}.
\label{subsec: projection assignment}
To avoid the aforementioned increased false negative problem, we can project the predicted $(x_i,y_i,w_i,h_i, \theta_i)$ as an axis-aligned box and then compute the overlap between projected box and axis-aligned ground truth. Then we have, 
\begin{align}
    \sigma_t^*(i) = \argmax_{j \in \braces{1,\ldots,m}} iou(P(x_i,y_i,w_i,h_i, \theta_i), P(B_j^t)), \\
    \tau_t^*(i) = \max_{j \in \braces{1,\ldots,m}} iou(P(x_i,y_i,w_i,h_i, \theta_i), P(B_j^t)).
\end{align}
where $P$ is the transformation defined in previous section. By doing this, we reduce the false negative rate as rotated boxes can be correctly assigned to the axis-aligned counterpart and increase the recall of the model. However, consider a rotated box that has a rotation angle of $\pi - \theta$ or $-\theta$. It will have the same projected axis-aligned box as the one with $\theta$ and these two are completely different boxes. Therefore, this process might misclassify negative examples as positive, which results in reduction in precision. The losses are:
\begin{align}
    L_{\mathcal{S}}^* &= \frac{1}{N} \sum_{i=1}^N -\mathbf{1}_{{\tau_s^*(i)} \geq 0.5}\log(p_i^*) + l_1(R_i^*, B_{\sigma(i)}^s), \\
    L_{\mathcal{T}}^* &= \frac{1}{N} \sum_{i=1}^N -\mathbf{1}_{{\tau_t^*(i)} \geq 0.5}\log(p_i^*) \\
    L^* &= L_{\mathcal{S}}^* + L_{\mathcal{T}}^*.
    \label{eq: loss rcnn}
\end{align}

\noindent\textbf{Heuristic Selection}.
Aside from the projection assignment, we also propose a parallel strategy. We use the same assignment rule as the source dataset~\ref{eq: rcnn source}. However, we evaluate how reliable each assignment is based on the aspect ratio and area size of axis-aligned ground truth and try to learn more from the reliable ground truth. For a particular axis-aligned ground truth box, we can compute its aspect ratio $r_j$ defined by $r_j = \max(w_j, h_j)/\min(w_j, h_j)$.
Larger aspect ratio implies the ground truth box is more reliable.
In addition, we can find an area threshold $a_{threshold}$ a minimum un-occluded stereotypical object area on a target dataset. For $w_j * h_j < a_{threshold}$, it is likely an occluded box. Following these heuristics, we define a binary reliability switch and loss for target dataset as
\begin{align}
g_i &= 
    \begin{cases}
        1 \text{ if ($r_{\sigma^*(i)} > 3$ or $w_{\sigma^*(i)} * h_{\sigma^*(i)} < a_{threshold}$})\\
        0 \text{ otherwise}
    \end{cases} \\
    L_{\mathcal{T}}^* &= \frac{1}{N} \sum_{i=1}^N - g_i\mathbf{1}_{{\tau_t^*(i)} \geq 0.5}log(p_i^*).
\end{align}
and we choose to mask unreliable examples. Despite heuristic selection is
less general than projection assignment, this can be beneficial for confined industry applications where the aspect ratio of the target object is known such as detecting a bottle on the conveyor belt.
\section{Experiments and Discussion}
\label{sec:results}
In this section, we first outline our implementation details and experimental setup including the datasets. Then we show the effectiveness of our method through  ablation studies followed by the comparison with different target and source datasets. Finally, we include some qualitative results.

\subsection{Experimental setup}
\noindent\textbf{Source Datasets.} Source datasets, in this work, are datasets strongly labelled with box orientations. The model has access to the source datasets to learn the task of rotated box detection. We choose a variety of source datasets including DOTA~\cite{xia2018dota}, COCO~\cite{lin2014coco} and Fresh-produce dataset~\ref{tab: dataset} where DOTA~\cite{xia2018dota} is a popular satellite imagery dataset with rotated annotations and COCO~\cite{lin2014coco} is a popular general object detection dataset. In COCO, rotated bounding box ground truth can be generated by finding the minimum enclosing rectangle on each instance segmentation mask ground truth. Fresh-produce dataset~\ref{tab: dataset} is a challenging dataset with high object density and heavy occlusion. The number of images and instances are shown in table~\ref{tab: dataset}. There are three long-shape subsets including banana, cucumber and carrot. 

\noindent\textbf{Target Datasets.} The training subset of the target datasets only contains axis-aligned bounding box ground truth which is used for the model to learn. Alternatively, the validation and test subsets contain rotated ground truth  to evaluate the performance of the model. We select HRSC2016~\cite{liu2016ship}, SSDD~\cite{SSDD2021}, cucumber and carrot datasets~\ref{tab: dataset} as our target datasets to cover a variety of domain permutations including satellite, single channel, natural and various object density.

\begin{table}
\small
  \caption{Fresh-produce datasets. }
  \label{sample-table}
  \centering
  \begin{tabular}{lcc}
    \toprule
    Class & Number of images & Number of instances\\ 
    \midrule 
    banana & 158 & 7391\\
    cucumber & 48 & 2036 \\
    carrot & 47 & 4647 \\
    \bottomrule
  \end{tabular}
  \label{tab: dataset}
\end{table}

\begin{table*}
\small
  \caption{Ablation Studies of the proposed methods. We use DOTA as source and fresh cucumber as target dataset. This table decoupled the effectiveness of rotated cotraining, rpn projection loss, single class, heuristic selection, projection assignment and axis-aligned pretraining using other objects.}
  \label{sample-table}
  \centering
  \begin{tabular}{ccccccc}
    \toprule
    Rotated Cotraining & RPN Proj. & Single Class & H. Select. & P. Assign. & Pretraining &{AP50$\uparrow$} \\
    \midrule 
          -     &     -      &      -     &      -     & -& -& 0.491 \\
     \checkmark &      -      &     -       &     -       & -&- & 0.542 \\
     \checkmark & \checkmark &      -      &       -     &- & -& 0.581 \\
     \checkmark & \checkmark & \checkmark &      -      & -& -& 0.633 \\
     \checkmark & \checkmark & \checkmark & \checkmark &    -        & -& 0.666 \\
     \checkmark & \checkmark & \checkmark &     -       & \checkmark & -& 0.664 \\
     \checkmark & \checkmark & \checkmark &        -    & \checkmark & \checkmark & \bf{0.683} \\
    \bottomrule
  \end{tabular}
  \label{tab: ablations}
\end{table*}

\begin{table}
\small
  \caption{Analysis for enlargement factor $\gamma$. We use DOTA as source and HRSC as target dataset for this analysis.}
  \begin{tabular}{cccccccc}
    \toprule
    $\gamma$ & 1.00 & 1.05 & 1.10 & 1.15 & 1.20 & 1.25\\
    {AP50$\uparrow$} & \textbf{0.791} & 0.763 & 0.756 & 0.733 & 0.718 & 0.650\\
    \bottomrule
  \end{tabular}
  \label{tab: gamma_ablation}
\end{table}

\noindent\textbf{Implementation Details.}
We pretrain our detector, Oriented-RCNN~\cite{xie2021oriented} using DOTA~\cite{xia2018dota}. The ship class and the images with the ships are removed from DOTA for this paper for pretraining and its role as a source dataset, due to class overlap with some target datasets. The main statistic we use in the paper to evaluate our method is AP50. It calculates the average precision with IOU threshold of $0.5$. Average precision (AP) is the area under precision and recall curve. 
We utilise the mmrotate framework~\cite{zhou2022mmrotate} and~\cite{xie2021oriented} for training. We use a batch of $2$ images from target dataset and a batch of $2$ images from the source dataset for a combined mini-batch of $4$. The losses and forward propagation for two batches are computed independently, then the losses are added and backpropagated through the network. We train the models up to $50$ epochs of target dataset. During inference, we set the non-maximum-suppression threshold to $0.5$ instead of $0.1$ used generally for aerial datasets, as the higher object density fresh-produce dataset contains far more object overlapping cases. We conduct the experiments using one 2080ti GPU with 11GB of memory. The training time depends on the dataset size, ranging from 30 minutes to 2 hours to train. The test can be done within 5 minutes for each dataset with speed of 15 FPS, which is the same as the original detector.

\subsection{Ablation Studies}

\begin{table}
\small
  \caption{We evaluate KCR with dense cucumber and carrot datasets with severe occlusion as target datasets. Tight rotated ground truth is used to evaluate AP50 performance of the model in the test time. We compare KCR method with original Oriented-RCNN~\cite{xie2021oriented} with weak axis-aligned annotation for training and fully-supervised models trained with strong rotated ground truth. Results show that KCR enables the model to perform on par with fully-supervised model and works reasonably well under large domain gap.}
  \label{tab:weak_compare}
  \centering
  \begin{tabular}{l|llcc}
    \toprule
    & Method            & Source & {Target Train} & {AP50$\uparrow$} \\
    
    \hline
    
    \parbox[t]{2mm}{\multirow{4}{*}{\rotatebox[origin=c]{90}{Cucumber}}} 
    
    & original~\cite{xie2021oriented}      & -              & axis-aligned       & 0.491             \\
    \cline{2-5}
    & KCR(ours)    & DOTA~\cite{xia2018dota}          & axis-aligned      & 0.683             \\
    & KCR(ours)    & banana         & axis-aligned      & \bf{0.788}        \\
    \cline{2-5}        
    & fully-supervised      & -              &rotated  & 0.820            \\
    
    \hline
    \hline
 \parbox[t]{2mm}{\multirow{4}{*}{\rotatebox[origin=c]{90}{Carrot}}}
    & original~\cite{xie2021oriented}      & -              & axis-aligned        & 0.560 \\
    \cline{2-5}
    & KCR(ours)    & DOTA~\cite{xia2018dota}          & axis-aligned        & 0.614 \\
    & KCR(ours)    & banana         & axis-aligned        & \bf{0.723}\\
    \cline{2-5}
    & fully-supervised      & -              & rotated  & 0.765\\
    \bottomrule
  \end{tabular}
\end{table}

\begin{table}
\small

 \caption{Evaluation of KCR method with HRSC2016~\cite{liu2016ship} and SSDD~\cite{SSDD2021} as target datasets against tight rotated ground truth in the test. Influence of source dataset is investigated using COCO~\cite{lin2014coco}, COCO with rotation augmentation and DOTA~\cite{xia2018dota}. We also compare KCR against original Oriented-RCNN~\cite{xie2021oriented}, Grabcut~\cite{rother2004grabcut} with weak axis-aligned annotation for training and fully-supervised model trained with strong rotated ground truth. We use projection assignment methods for KCR except for the ones with HS, which stands for heuristic selection.}
  \label{tab: hrsc_ssdd}
  \centering
  \begin{tabular}{l|llcc}
  \toprule
    & Method            & Source & {Target Train} & {AP50$\uparrow$} \\

    \hline
    
    \parbox[t]{2mm}{\multirow{7}{*}{\rotatebox[origin=c]{90}{HRSC2016~\cite{liu2016ship}}}} 
    
    & original~\cite{xie2021oriented}  & -        & axis-aligned     & 0.175            \\
    \cline{2-5}
    & Grabcut\cite{rother2004grabcut}           & -                & axis-aligned             & 0.240            \\
    & Grabcut Train           & -                & axis-aligned            & 0.629            \\
    & KCR(ours)    & COCO~\cite{lin2014coco}             & axis-aligned             & 0.579            \\
    & KCR(ours)    & COCO Aug      & axis-aligned             & 0.783         \\     
    & KCR(ours)    & DOTA~\cite{xia2018dota}           & axis-aligned             & \bf{0.791}            \\
    & KCR-HS(ours)    & DOTA~\cite{xia2018dota}           & axis-aligned             & \bf{0.778}            \\
    \cline{2-5}
    & fully-supervised   & -                & rotated            & 0.903            \\ 
    \hline
    \hline
    \parbox[t]{2mm}{\multirow{6}{*}{\rotatebox[origin=c]{90}{SSDD~\cite{SSDD2021}}}} 
    & original~\cite{xie2021oriented}  & -                & axis-aligned             & 0.432            \\
    \cline{2-5}
    & Grabcut Train          & -                & axis-aligned             & 0.585            \\
    & KCR(Ours)    & COCO~\cite{lin2014coco}              & axis-aligned             & \bf{0.888}               \\
    & KCR(Ours)    & COCO Aug    & axis-aligned             & 0.874              \\
    & KCR(Ours)    & DOTA~\cite{xia2018dota}            & axis-aligned             & 0.881           \\
    & KCR-HS(Ours)    & DOTA~\cite{xia2018dota}            & axis-aligned             & 0.874           \\
    \cline{2-5}
    & fully-supervised      & -                & rotated           & 0.898         \\
    \hline
  \end{tabular}
\end{table}

To tackle the weakly-supervised learning of rotated bounding box given only axis-aligned ground truth, we build our approach progressively. In this section, we show how the approach evolves using cucumber as our axis-aligned training target dataset and DOTA~\cite{xia2018dota} as our source dataset. We choose this particular pair because their domain gap is large in terms object appearance, density and occlusion severity. It is more convincing if the model is able to learn rotation with such a large domain gap between source and target.

We first establish a baseline by training a rotated detector using our axis-aligned cucumber training dataset and test it on rotated cucumber test set. This baseline is the first row of table~\ref{tab: ablations} with AP50 of 0.491.
Then we introduce the cotraining strategy with the source dataset without any modification of the training scheme. That means we simply treat both source and target datasets equally and completely follow the training scheme of ~\cite{xie2021oriented}. This gives us AP50 of 0.542 which is an incremental improvement over the baseline. 

RPN projection~\ref{eq: loss rpn} improves the performance to 0.581 (Table~\ref{tab: ablations}). Instead of learning the wrong rotation supervision from the axis-aligned dataset, it chooses not to learn. After that, we try the single class strategy: only use one class label for all classes in source dataset. This effectively improves the performance to 0.633 because the model focuses on learning the object score and rotation angle instead of the classification problem of the source dataset. If the target dataset is multi-class, we can apply the single class strategy to the source dataset because the classification is not important. 

The projection assignment and heuristic selection methods are implemented separately. The projection assignment strategy as described in section~\ref{subsec: Learning Rotated R-CNN} improves the performance to 0.664 since it reduces the false negative rate. It cannot further improve performance due to the false positive assignment problem. Heuristic selection strategy improves the model performance to 0.666. Finally, we pretrain this model on other axis-aligned objects and improve this performance to 0.683 which is 0.192 higher than the baseline. We take the last row of table~\ref{tab: ablations} as our final approach and apply it to permutations of target and source datasets in section~\ref{sec: different asin}.

We also provide an ablation of the enlargement factor $\gamma  \geq 1$. As we can see from Table~\ref{tab: gamma_ablation}, $\gamma = 1.00$ is the best, which is what we have used throughout all the experiments. 

\begin{figure*}
  \centering
  \includegraphics[width=0.95\linewidth]{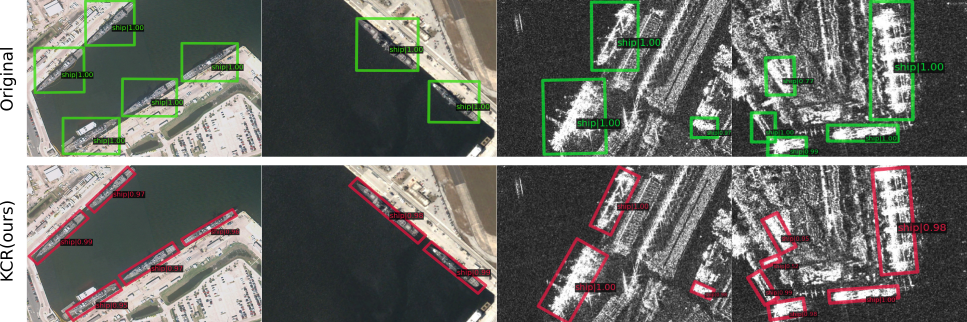}
  \caption{Visualization of KCR performance against original Oriented-Rcnn trained with weak axis-aligned annotation. The images are from test sets of HRSC2016~\cite{liu2016ship} and SSDD~\cite{SSDD2021}. We use the COCO as our source dataset which has large domain gap from the target. The model trained with KCR methods learnt to predict accurate rotated bounding box, which is much more precise than the original model. }
  \label{Figure: quali_ship}
  \vspace{-1em}
\end{figure*}

\begin{figure}[t]
    \centering
    \includegraphics[width=0.9\linewidth]{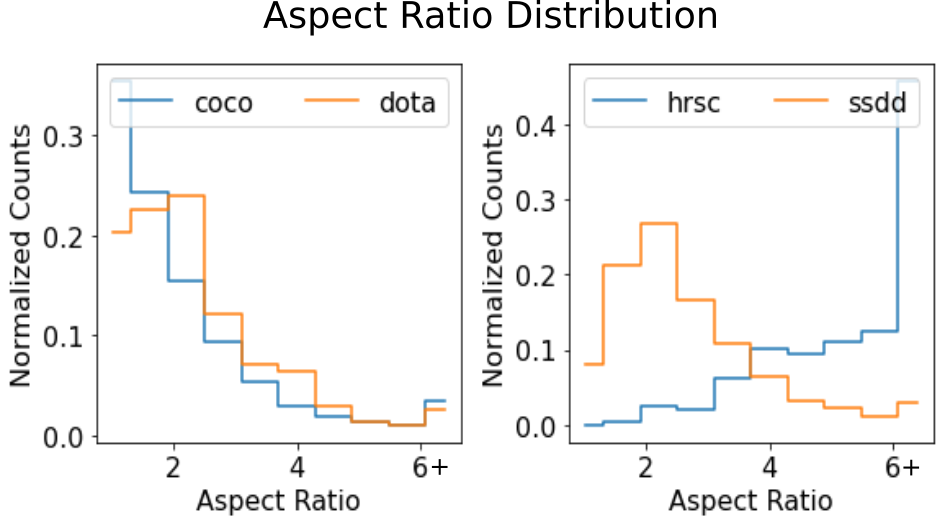}
    \caption{Distribution of Aspect ratios on 4 datasets.}
    \label{fig:asp} 
    \vspace{-2em}
\end{figure}  

\begin{figure*}
  \centering
  \includegraphics[width=0.85\linewidth]{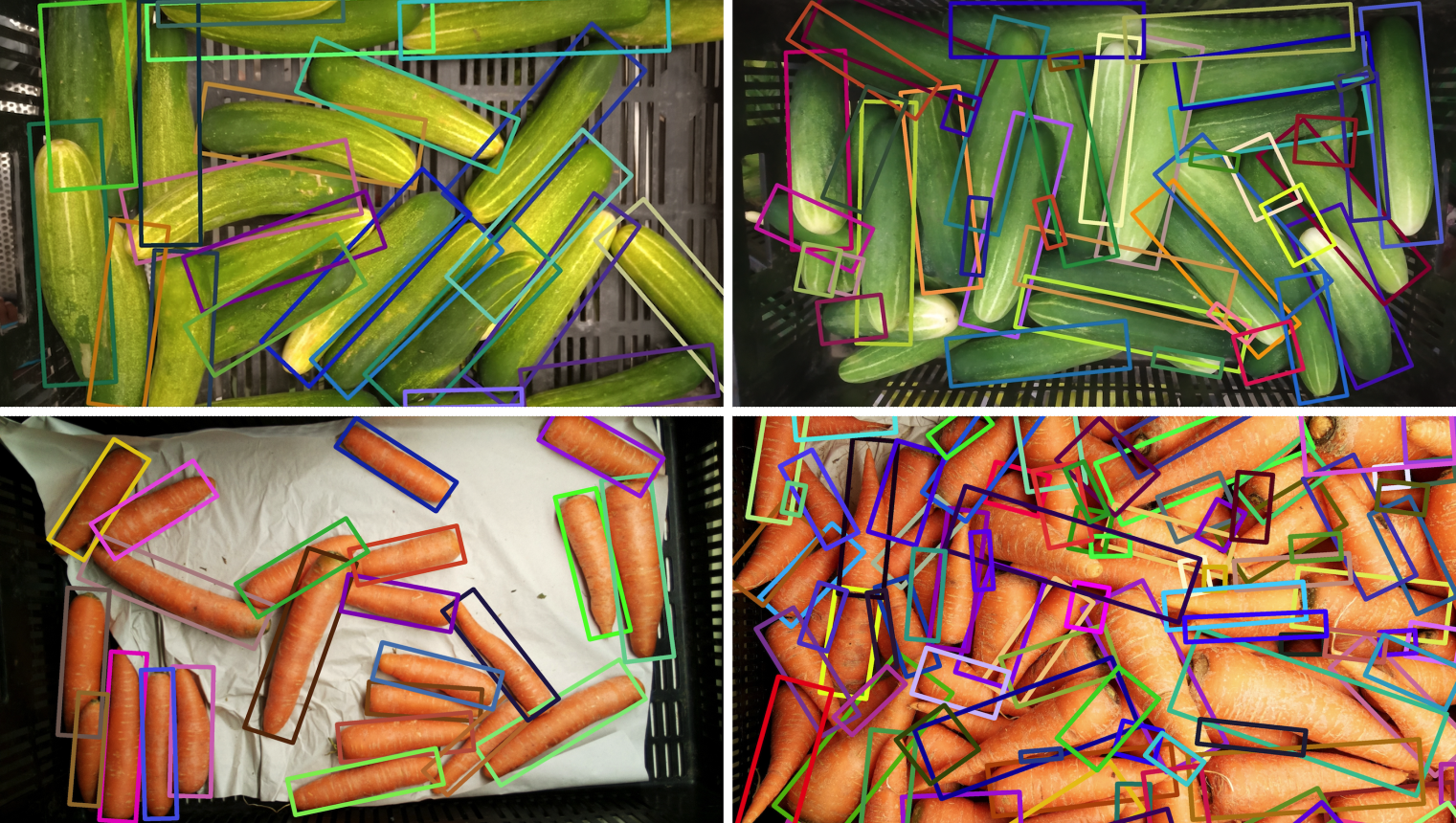}
  \caption{Visualization of KCR performance trained with axis-aligned only dataset. The images are chosen from an unlabeled set to ensure generality. The model is clearly capable of predicting tight rotated bounding box even with high object density and occlusion.}
  \label{Figure: quali}
\end{figure*}

\label{sec: different asin}

\subsection{Main results}
In this section, we apply KCR framework to various combinations of source and target datasets to investigate the generality of the method under different domain gaps. We also compare KCR with a popular foreground segmentor~\cite{rother2004grabcut}.

\noindent\textbf{Fresh-produce dataset}.
In Table \ref{tab:weak_compare} we can see results progressively improve when information that is closer to desired result is introduced. We initially evaluate our method where no knowledge combination is used, that is, for the cucumber and carrot, we use the axis-aligned bounding boxes and allow the network to learn incorrect $\alpha, \beta$ and $\theta$ values. In the next iteration, we use DOTA\cite{xia2018dota} aerial satellite images for knowledge combination. The domain gap between these images and the cucumber and carrots is large because of the sparsity of objects in DOTA. Nonetheless, 
introducing rotated-bounding box information, even with dissimilar dataset, results in a large performance improvement for both cucumber and carrot as seen in the second row of Table \ref{tab:weak_compare}. The next progression combines knowledge with bananas annotated with rotated bounding-box. This further improves the results and provides an insight into task similarity strengthening knowledge combination. The banana dataset offers the opportunity for the model to learn rotated detection with severe occlusion. For reference, the last row in Table \ref{tab:weak_compare} is the result when training the model directly on ground truth rotated bounding box, i.e. fully supervised training. We note that the performance gap between our approach and fully supervised training is small.

\noindent\textbf{HRSC2016}. HRSC2016~\cite{liu2016ship} is a rotated satellite imagery dataset that focuses on ships. The objects in HRSC2016 typically have a larger aspect ratio and occupies larger area of the image in comparison DOTA~\cite{xia2018dota} targets. To prevent class overlap, images with ships have been removed from DOTA when using as a source dataset. We establish our baseline by using original training regime and only axis-aligned ground truth of HRSC2016 yielding AP50 of 0.175, which is 72.8\% lower than a fully supervised result as shown in Table~\ref{tab: hrsc_ssdd}. We investigate the performance of KCR with COCO~\cite{lin2014coco}, containing only natural images, as source dataset because the domain gap between COCO and HRSC2016 is visually large. We generate rotated bounding box ground truth from instance segmentation masks. The result significantly improves to 0.579 when we use KCR framework. Although the rotated bounding box is used in source dataset, objects in COCO are typically axis-aligned, hence produce a weak rotation training signal. We therefore rotate images by $0-180$ degrees followed by horizontal and diagonal flip augmentation to increase the number of rotated examples in the source dataset. As a result, AP50 rises to 0.783 with strongly augmented COCO. Thus we show that KCR enables the transfer learning of extra parameter under large domain gap. At last, we use DOTA dataset as source and achieves AP50 of 0.791, which is only 11.2\% lower than the fully supervised model. The performance gap is potentially due to lack of equally long objects with large aspect ratio in the source dataset.

\noindent\textbf{SSDD}. SSDD~\cite{SSDD2021} is SAR dataset which also focuses on detection of ships. Images in SSDD are single channel and vary in frequency depending on the sensor used for acquisition. These images are commonly low resolution and contain high frequency noise, which results in a visually large domain gap from any source datasets. We follow the same knowledge combination strategy as before. Using KCR, a trained rotated box model performs almost equally well from three different source datasets (DOTA, coco and coco augmented). The final AP50 of 0.888 is 45.6\% better than the baseline and only 1\% lower than a fully supervised model.

\noindent\textbf{Rotated Boxes with Grabcut.}
GrabCut~\cite{rother2004grabcut} is a computer vision algorithm that predicts a foreground initialised with a region-of-interest. This foreground segmentation can then be used for predicting a rotated box with a variety of heuristic algorithms. This predicted rotation can be used at inference time as a post-processing step, transforming the axis aligned boundary box to a rotated boundary box. At test time, this improves the baseline by a minimal 6.5\%, however reduces inference throughput to 0.3 FPS. Alternatively, this algorithm can be performed offline on the axis-aligned ground truth to produce a noisy rotated boundary box ground truth for training. This method is significantly more effective, improving test time performance on rotated ground truth to 0.629 AP50 for HRSC and 0.585 for SSDD. However, KCR outperforms training on noisy ground truth by 16.2\% on HRSC and 30.3\% on SSDD.

\noindent\textbf{Analysis on aspect ratios.} We show histograms of aspect ratio in Fig~\ref{fig:asp}. The biggest difference between an axis-aligned and rotated box happens when the instance has high aspect ratio \textbf{and} rotated. Most COCO objects are neither long \textbf{nor} rotated while DOTA has a fair distribution of long and rotated objects. 
Performance gap between original COCO and DOTA as source dataset is bigger for HRSC than SSDD because HRSC has a more tail heavy aspect ratio distribution.

\subsection{Qualitative Results}
We visualize performance of KCR on HRSC2016~\cite{liu2016ship} and SSDD~\cite{SSDD2021} in Figure~\ref{Figure: quali_ship}. The model has successfully learnt to predict accurate rotated bounding box with weak axis-aligned annotation. The source dataset we use to gain rotation knowledge is COCO~\cite{lin2014coco}, which has a large domain gap distant from the target dataset. The rotated prediction from our framework produces boxes much tighter than a model which was trained on and predicts axis-aligned boxes.

Fresh-produce dataset is more challenging due to higher density of objects with severe occlusion. We visualize the performance of KCR with cucumber and carrot as target datasets and a banana dataset as the source dataset supplying knowledge of rotation. The depicted images are from a separate unlabeled set to ensure generality. As shown in Figure~\ref{Figure: quali}, the model is clearly capable of predicting tight rotated bounding boxes in a challenging scenario which is core contribution of this paper. The model is able to complete the task with high precision and recall in a scene with frequent and extremely occlusions, various lighting conditions and different object sizes. For the two images in the left column, the model detects almost every object.

\section{Conclusion}
Rotated detection improves the performance of downstream tasks by reducing the overall area of the enclosing box, improving the foreground to background ratio. This is particularly important for scenes with high density of targets and complex occlusions. However, most existing datasets only provide axis-aligned annotation, with the lack of the capability to annotate rotated boxes. In this paper, we address this problem by proposing KCR, which is a novel knowledge combination training scheme that only requires axis-aligned annotation for the target object class to train the model. At inference time, the model predicts accurate rotated bounding boxes on par with fully-supervised approach. This approach will enable the detector to predict an extra but crucial parameter. We believe this work can greatly extend the use case of rotated object detection by reducing annotation costs.

{

\bibliographystyle{plain}
\bibliography{ref}
}

\end{document}